\documentclass[10pt,twocolumn,letterpaper]{article}

\usepackage{lineno}
\usepackage[pagenumbers]{iccv} 

\definecolor{iccvblue}{rgb}{0.21,0.49,0.74}
\usepackage[pagebackref,breaklinks,colorlinks,allcolors=iccvblue]{hyperref}
\usepackage{adjustbox}
\usepackage{diagbox}
\usepackage{graphicx}
\usepackage{amsmath}
\usepackage{amssymb}
\usepackage{booktabs}
\usepackage[dvipsnames]{xcolor}
\usepackage{flexisym}
\usepackage{amsmath}
\usepackage{multirow}
\usepackage{tabularx}
\usepackage{enumitem}

\usepackage{subcaption}
\usepackage{tikz}

\usepackage{amssymb}
\usepackage{pifont}

\newcommand{\NA}[1]{$\spadesuit$\footnote{\color{red}: #1}}

\def\paperID{655} 
\def\confName{ICCV}
\def\confYear{2025}

\title{DynaGSLAM: Real-Time Gaussian-Splatting SLAM for Online Rendering, Tracking, Motion Predictions of Moving Objects in Dynamic Scenes}

\author{Runfa Blark Li\textsuperscript{*}\textsuperscript{$\dagger$} \quad
Mahdi Shaghaghi\textsuperscript{$\dagger$} \quad
Keito Suzuki\textsuperscript{*} \quad
Xinshuang Liu\textsuperscript{*} \quad
Varun Moparthi\textsuperscript{*} \quad \\
Bang Du\textsuperscript{*} \quad 
Walker Curtis\textsuperscript{$\dagger$} \quad
Martin Renschler\textsuperscript{$\dagger$} \quad 
Ki Myung Brian Lee\textsuperscript{*} \quad \\
Nikolay Atanasov\textsuperscript{*} \quad 
Truong Nguyen\textsuperscript{*} \quad
\\ [0.5em]
\textsuperscript{*}UC San Diego \\
\textsuperscript{$\dagger$}Qualcomm XR Advanced Technology \\
{\tt\small \{runfa, k3suzuki, xil235, vmoparthi, b7du, kmblee, natanaso, tqn001\}@ucsd.edu} \\
{\tt\small \{shaghagh, wcurtis\}@qti.qualcomm.com} \\
\url{https://blarklee.github.io/dynagslam/}
}

\begin{document}
\twocolumn[{%
\renewcommand\twocolumn[1][]{#1}%
\maketitle
\begin{center}
\begin{minipage}[t]{\textwidth}
\centering
\includegraphics[width=\linewidth]{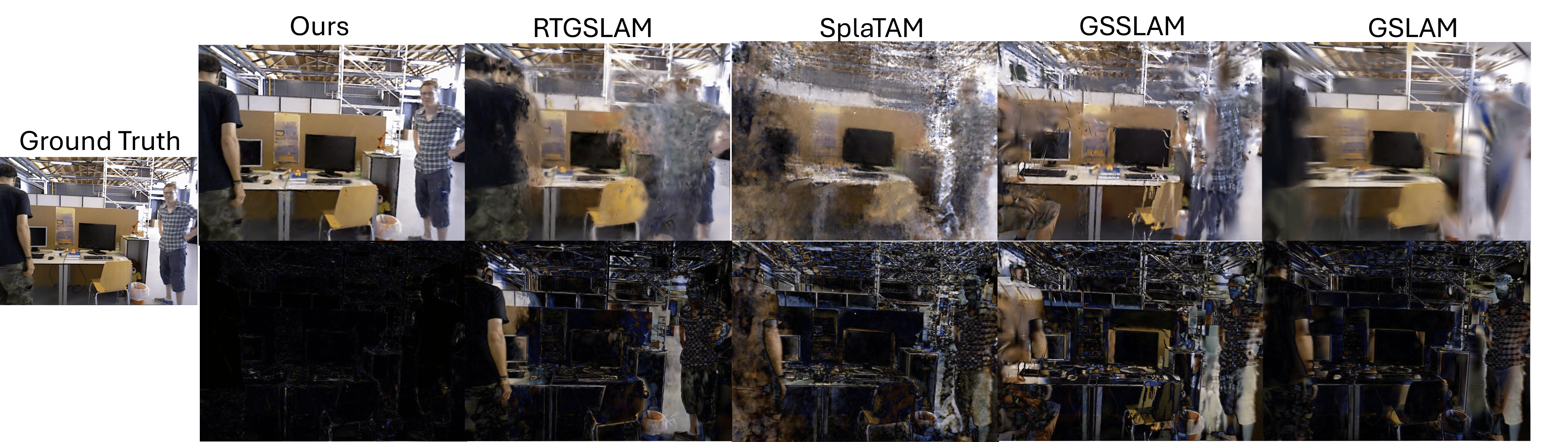}
\end{minipage}
\vspace{-5mm}
\captionof{figure}{\textbf{DynaGSLAM} is the first real-time Gaussian-Splatting (GS) based SLAM for online high-quality rendering of dynamic objects (moving people in this scene) in dynamic scenes, while jointly estimating ego motion. With the online RGBD frames, DynaGSLAM enables us to track(interpolate)/predict(extrapolate) the continuous object motions in the past/future. This figure shows the rendering of GS mapping on TUM dataset \cite{tum} \protect\footnotemark. First row: RGB rendering. Second row: Absolute error between the rendering and the ground truth.}
\label{fig:fig1}
\end{center}
}]
\footnotetext{TUM RGB-D SLAM dataset and benchmark is licensed under Creative Commons 4.0 Attribution License (CC BY 4.0).}

\begin{abstract}
Simultaneous Localization and Mapping (SLAM) is one of the most important environment-perception and navigation algorithms for computer vision, robotics, and autonomous cars/drones. Hence, high quality and fast mapping becomes a fundamental problem. With the advent of 3D Gaussian
Splatting (3DGS) as an explicit representation with excellent rendering quality and speed, state-of-the-art (SOTA) works introduce GS to SLAM. Compared to classical pointcloud-SLAM, GS-SLAM generates photometric information by learning from input camera views and synthesize unseen views with high-quality textures. However, these GS-SLAM fail when moving objects occupy the scene that violate the static assumption of bundle adjustment. The failed updates of moving GS affects the static GS and contaminates the full map over long frames. Although some efforts have been made by concurrent works to consider moving objects for GS-SLAM, they simply detect and remove the moving regions from GS rendering (``anti'' dynamic GS-SLAM), where only the static background could benefit from GS. To this end, we propose the first real-time GS-SLAM, ``DynaGSLAM'', that achieves high-quality online GS rendering, tracking, motion predictions of moving objects in dynamic scenes while jointly estimating accurate ego motion. Our DynaGSLAM outperforms SOTA static \& ``Anti'' dynamic GS-SLAM on three dynamic real datasets, while keeping speed and memory efficiency in practice.
\end{abstract}    
\section{Introduction}
\label{sec:intro}

\begin{figure}[t]
    \centering
    \includegraphics[width=\linewidth]{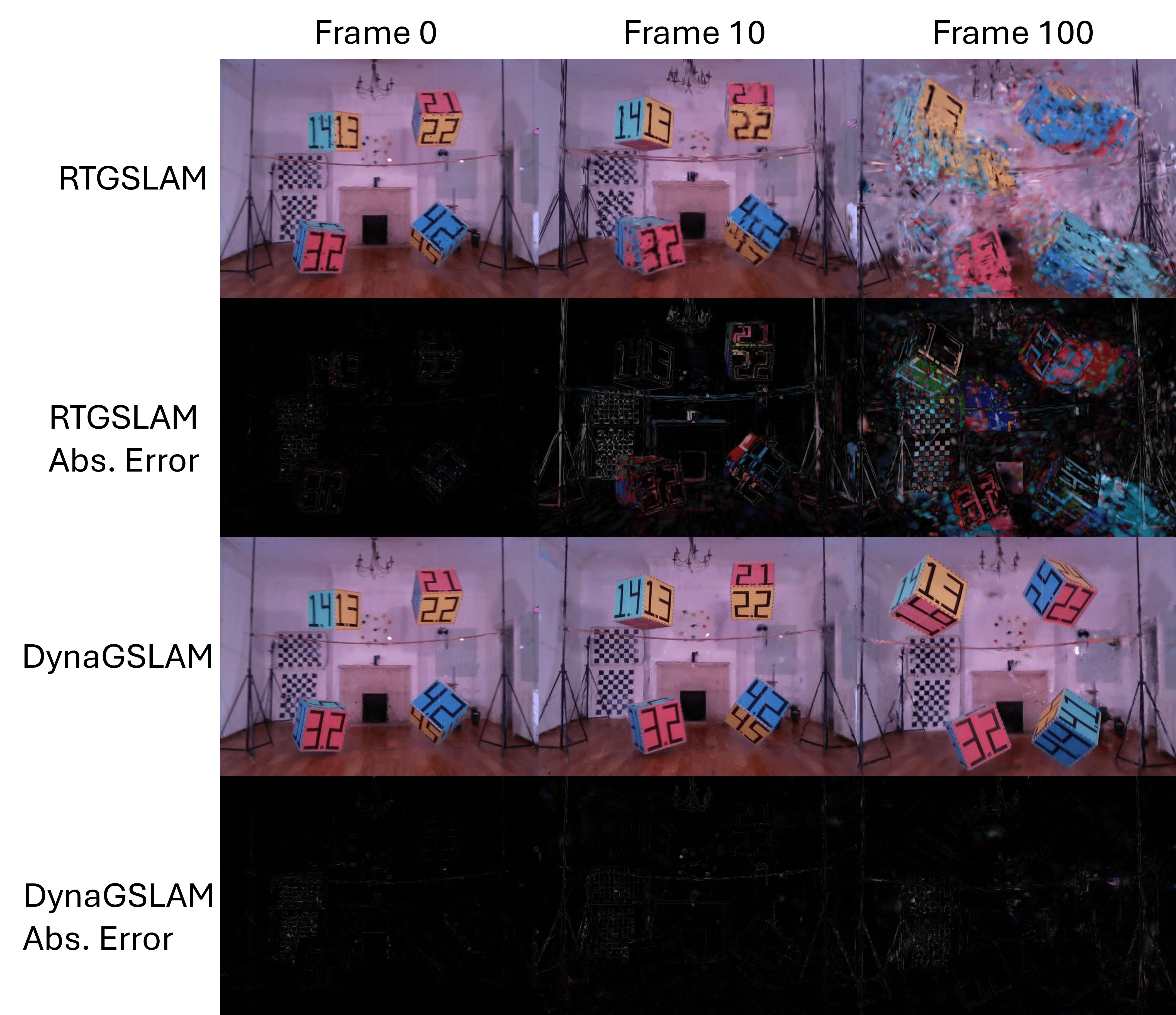}
    \vspace{-5mm}
    \caption{\textbf{Failures of static GS-SLAM over long frames.} ``Avs. Error'' is the absolute error of the rendered RGB to the ground truth RGB. Typical static GS-SLAM works (like RTGSLAM \cite{rtgslam}) do not consider moving objects, not only the rendering quality of the moving objects becomes worse over long frames, but the static regions get contaminated by the failed dynamic GS. Our DynaGSLAM consistently renders both dynamic and static regions in high quality over long frames.} 
    \label{fig:fig2}
\end{figure}

Simultaneous localization and mapping (SLAM) is a fundamental problem with wide areas of application in robotics, virtual reality, and autonomous vehicles.
Given an input RGBD video sequence, a SLAM algorithm must jointly estimate the ego motion (camera pose) and build the map of the environment, conventionally represented as a point cloud. 
Although existing SLAM algorithms are robust and real-time in simple cases \cite{orbslam2}, there remain two major limitations that cannot be neglected: 1. Standard SLAM assumes a static scene, and cannot gracefully handle dynamic objects; 2. The pointcloud representation conveys basic 3D structure of the environment, but offers no photometric information on graphics or texture other than the input RGB, limiting downstream tasks. 

Many efforts \cite{nice-slam, orbslam3, dynavins, mvo, clusterslam, vdoslam, dynosam} have been made to improve SLAM with moving objects. 
While implementations vary, the main idea is to detect and remove the moving objects that violate static assumptions for bundle adjustment. Some SOTA works \cite{dynosam, vdoslam} take a step further to track object motions. This may be at the expense of computational speed; however, explicit consideration of moving objects improve robustness in many challenging indoor/outdoor scenes. Nonetheless, the pointcloud representation still offers limited photometric information.

With the rise of Gaussian splats (GS) \cite{3dgs} as an explicit 3D representation with effective rendering speed and quality, recent work
adopted GS to capture photometric information in SLAM~\cite{gsslam0,gsslam_cvpr1,gsslam_cvpr2,mm3dgsslam,photoslam,rp-slam,splatam,rtgslam}. 
This body of work replaces the conventional pointclouds with GS, so that RGB images from novel views can be synthesized beyond the input camera views using GS. 
Such a representation retains the 3D structure with explicit GS, while adding significant photometric information. 
However, these SOTA methods only consider with static scenes.
Therefore, as shown in Fig. \ref{fig:fig2}, even static regions are corrupted by the mishandling of dynamic regions. 
A few concurrent methods \cite{dgslam,dgsslam,garadslam,gassidy,pgslam} realize "anti" dynamic SLAM with GS representation, where dynamic objects are detected and removed.
Although this benefits localization, removing dynamic objects means only the static background can be rendered from the GS map.

To represent dynamic objects with GS, SOTA methods \cite{dynamic3dgs,4DGS,deformable3dgs,swinggs,spacetimegs,3dgstream,gaufre,realtime-4dgs,4drotorgs,egogaussian,compact-3DGS,dynmf,factor-field,fullyexplicitdynamicgaussian,splineGS,mosca} explore directly adding a time dimension to GS.
However, these methods train GS in an offline manner, for hours per video sequence, and are hence unsuitable for online SLAM. 
In contrast to all concurrent SOTA GS-SLAM or dynamic GS, our contributions can be summarized as:
\begin{itemize}
\item We propose DynaGSLAM, the first real-time Gaussian-Splatting based SLAM that achieves high-quality online GS rendering, tracking, motion predictions of moving objects with dynamic ego motions in the scene.  
\item We propose a novel dynamic GS management algorithm for adding, deleting, tracking, updating and predicting dynamic GS, specifically our novel online GS tracker and the motion interpolation/extrapolation algorithms, which enable online real-time accurate dynamic GS mapping with humble memory requirements.
\end{itemize}

\section{Related Works}
\noindent
\textbf{Dynamic SLAM.}
To handle dynamic objects, the usual approach is to detect and subtract the dynamic regions of the image. Earlier work \cite{orbslam3,dynavins} use RANSAC \cite{ransac} or point correlations \cite{point_correlation} for motion detection, and the recent learning-based methods \cite{dynaslam, flowfusion} learn to semantically segment moving objects. 
These methods improve the quality of camera pose estimation by removing dynamic objects, but lose the object motion information. 
Hence, we refer to these methods as ``anti'' dynamic SLAM. 
To extract the object motion, some dynamic SLAM incorporate and track the dynamic objects. 
These methods assume the objects are rigid, and assign a tracklet to every object.
DynaSLAM2 \cite{dynaslam2} tracks rigid objects by estimating the motion of centroids, which also improves camera localization. 
SOTA work DynoSAM \cite{dynosam} proposes a world-centric factor-graph optimization for accurate but suffers from time-consuming object motion estimation. 
However, all these attempts are based on classical point cloud mapping, it is non-trivial to directly extend the ideas to GS-SLAM since GS have more complex attributes, to be optimized other than the point position, such as spherical harmonic and the shape (covariance).

\noindent
\textbf{GS-based SLAM.}
GS \cite{3dgs} has become a promising alternative to point cloud for SLAM \cite{gsslam0,gsslam_cvpr1,gsslam_cvpr2,splatam,rtgslam}. 
Initialized from an RGBD point cloud, new GS are incrementally added with ego motion to complete the scene. 
Compared to point cloud, 3DGS contains high-quality photometric information, at the expense of additional storage and computation. 
Thus, SOTA GS-SLAM focus on GS management algorithms, where RTGSLAM \cite{rtgslam} designed a representative real-time algorithm for the tasks of adding, deleting, and reusing of static GS by converting GS in stable and unstable status.
However, these SOTA GS-SLAM methods only work for static scenes.

There are concurrent efforts on extending GS-SLAM to dynamic scenes \cite{dgslam,dgsslam,garadslam,gassidy,pgslam}.
However, these methods fall under the ``anti'' dynamic SLAM category because they segment and discard dynamic objects. 
Our method, DynaGSLAM, is the first to construct a dynamic GS that models dynamic objects in the online SLAM setting.


\noindent
\textbf{Offline Dynamic GS.} Dynamic GS \cite{dynamic3dgs,4DGS,deformable3dgs,swinggs,spacetimegs,3dgstream,gaufre,realtime-4dgs,4drotorgs,egogaussian,compact-3DGS,dynmf,factor-field,fullyexplicitdynamicgaussian,splineGS,mosca} has been also attracting lots of attention. With video and camera poses over frames, dynamic GS aims to train GS in ``4D'' such that the well-trained GS can be rendered from unseen views at any given timestamps in the video. \cite{4drotorgs, realtime-4dgs} explored an explicit additional time dimension for GS position and shape, and design 4D rotation matrix with 4D-rotor and extend Spherical Harmonics to 4D, but the extra time dimension on all GS introduce speed and memory burden. The motion-function based dynamic GS \cite{gaussian-flow,compact-3DGS,dynmf} leveraged Fourier series\& cubic polynomials for translation and SLERP (Spherical Linear Interpolation) for rotation, and embed the motion function parameters as GS attributes. However, these strategies introduce additional channels to all Gaussians, and require accurate supervision over time to learn the motion. Some methods can also segment moving GS \cite{gaufre,egogaussian,motion-aware-gs,fullyexplicitdynamicgaussian}, but the motion-awareness is only achieved when the dynamic GS is fully trained over all frames. Worse still, these methods all require long, offline training, and are not suitable for online SLAM.
Compared to the offline dynamic GS methods, our dynamic GS-SLAM method performs equally well with three challenging constraints: 1) the target images are presented \emph{online}, so future frames are inaccessible; 2) GS is optimized in real-time, while capturing dynamic objects; and 3) the camera trajectory is unknown or inaccurate. 



\begin{figure*}[!t]
    \centering
    \includegraphics[width=\linewidth]{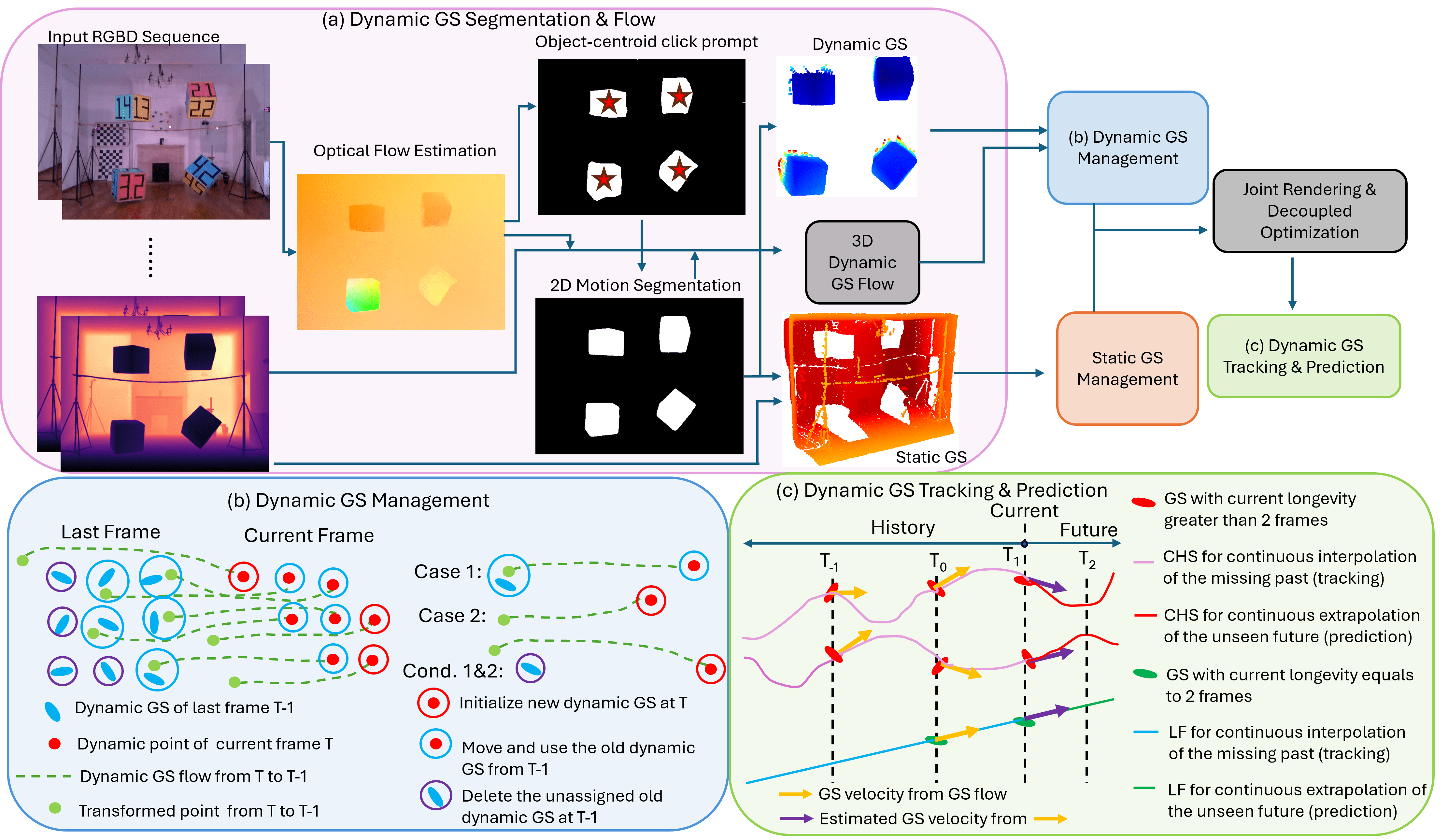}
    \vspace{-5mm}
    \caption{\textbf{Overview of DynaGSLAM Mapping.} We focus on three modules - Dynamic GS (a) Segmentation \& Flow, (b) Management and (c) Tracking \& Prediction. DynaGSLAM takes RGBD sequence as input to construct map with GS, (a) segment dynamic GS from static GS in 3D, and estimate dynamic GS 3D motion flow between frames. (b) Dynamic GS are managed separately from static GS with GS flow, but combined to jointly optimize. Case 1 \&2 are the rules for dynamic GS adding; ``Cond. 1\&2'' denotes the conditions for dynamic GS deletion. (c) The optimized dynamic GS at current and past frames are used to interpolate/extrapolate dynamic GS in the continuous timeline from past to future. ``CHS'' refers to ``cubic Hermite spline'' and ``LF'' refers to ``linear function''. Localization details are not included in the figure since our main contribution is on mapping.} 
    \label{fig:overview}
\end{figure*}

\noindent


\section{Problem Formulation}\label{sec:problem_formulation}
We formulate the dynamic GS SLAM problem as follows.
We are given a streaming sequence of RGB and depth images $C_t\in \mathbb{R}^{W\times H \times 3}$ and $D_t \in \mathbb{R}^{W\times H}$ of a scene, taken from unknown camera poses $T_{t}\in SE(3)$.
The scene contains moving objects.

The objective is to recover the unknown camera poses $T_{t}$ (i.e., localization) and to find a time-varying scene representation $\mathcal{G}_{t}$ that models the moving objects (i.e., mapping). We use the GS to represent $\mathcal{G}_{t}$, so that RGB and depth images $\hat{C}(\mathcal{G}_t, T_t)$ and $\hat{D}(\mathcal{G}_t, T_t)$ can be synthesized to match the scene at time $t$ seen from camera pose $T_t$. To this end, at each time $t$, we aim to find the camera trajectory $\{T_{\tau}\}$ for $\tau \in [0, t]$, and a time-varying scene representation $\{\mathcal{G}_\tau\}$, such that:
\begin{equation}
\min_{\mathcal{G}_\tau, T_{\tau}} \sum_{\tau=0}^{t} \ell_c(\hat{C}(\mathcal{G}_\tau,T_\tau),C_\tau) + \ell_d(\hat{D}(\mathcal{G}_\tau,T_\tau),D_\tau)
\end{equation}
where $\ell_c$ and $\ell_d$ are color and depth image losses measuring the similarity between images $\hat{C}_t, \hat{D}_t$ reconstructed by GS and the images $C_t$, $D_t$ provided by the camera. Furthermore, we focus on tracking a time-varying GS $\mathcal{G}_t$ over a time horizon, rather than only creating new GS for each timestep.
This allows photorealistic synthesis of images not only at novel viewpoints (as in static GS), but also at \emph{continuous} novel times.
For simplicity, our notations reflect the case of regular, unit time intervals; however, we aim to predict and track motion over continuous time, given data arriving at irregular time intervals. 

To ensure online, real-time performance, we treat the localization and mapping problems separately. 
For localization, we rely on DynoSAM~\cite{dynosam}, which is a SOTA graph-based visual localization method in dynamic scenes based on point cloud representation.
Using the camera trajectory $T_t$ estimated by DynoSAM, we focus on the mapping problem of finding the dynamic GS $\mathcal{G}_{t}$, which allows photorealistic image rendering that is not possible with point clouds.
For example, this is analogous to the use of ORB-SLAM2 \cite{orbslam2} in RTGSLAM \cite{rtgslam}. We defer the correction of camera trajectory using the dynamic GS $\mathcal{G}_t$ to future work, with the expectation that the improvements will be incremental.

\section{Dynamic GS Architecture}



To solve the problem of photorealistic synthesis of dynamic objects, we introduce a new variant of GS with a dynamic mean that moves over time.
We define a dynamic GS as a set of Gaussian blobs defined as $\mathcal{G}_{t} = \left\{ (\mathbf{m}_{t}^i(\tau), \Sigma_{t}^i, \alpha_{t}^i, \mathbf{sh}_{t}^i) \right\}$
, where $\Sigma_{t}^i \in \mathbb{R}^{3 \times 3}$, $\alpha_{t}^i \in \mathbb{R}$ and $\mathbf{sh}_{t}^i \in \mathbb{R}^{16}$ are the covariance matrix, opacity and spherical harmonics. Importantly, $\mathbf{m}_{t}^i(\tau)$ is a time-varying mean allowing novel-view synthesis at an unobserved time $\tau$, modeled as a cubic Hermite spline:
\begin{equation}
\begin{aligned}
\mathbf{m}_{t}^{i}(\tau) &= (2{\tau'}^{3} - 3{\tau'}^{2} + 1) \mathbf{m}_{t-}^{i} + ({\tau'}^{3} - 2{\tau'}^{2} + {\tau'}) \mathbf{v}_{t-}^{i} \\
              &\quad + (-2{\tau'}^3 + 3{\tau'}^2) \mathbf{m}_{t+}^i + ({\tau'}^3 - {\tau'}^2) \mathbf{v}_{t+}^i,
\end{aligned}
\label{eq:cubic_hermite}
\end{equation}
where $\tau' = \tau - t - 1$, $\mathbf{m}_{t-,t+}^i$ and $\mathbf{v}_{t-,t+}^i$ are interpolation parameters. Notably, these parameters can be updated analytically without iterative optimization, as we detail later.
Extrapolation into the future is achieved by querying $\tau > t$. 

RGB images are rendered similarly to original 3DGS~\cite{3dgs} through alpha-blending of projected Gaussians at each time $t$:
\begin{equation}
\begin{split}
\hat{C}_{t} &= \sum_i c_t^i f(g_t^i) \prod_{j=1}^{i-1} ( 1 - f(g_t^j)),
\end{split}
\label{eq:volumetric_rendering}
\end{equation}
where $f(g_t^i) = \alpha_t^i \mathcal{N}_{\text{2D}}(P\mathbf{m}_t^i(t), P\Sigma_t^iP^{T} )$ is the weight of the $i$-th Gaussian at time $t$, after affine projection $P$.




For rendering accurate depth images $\hat{D}_t$, we adopt the ideas from 2DGS \cite{2DGS}, and discard the shortest principal axis of 3D Gaussians for higher efficiency and better surface representation. 
We also adopt the ``surface rendering'' technique for depth from~\cite{2DGS} as it is faster than alpha blending. This takes the depth $d(g_{t}^{i}$) of the closest GS that is over an opacity threshold $\lambda_{\alpha}$:
%
\begin{equation}
\hat{D} = \min_{g_{t}^i \in \mathcal{G}_{t} } d(g_{t}^i), \text{ s.t. } f(g_t^i) \prod_{j=1}^{i-1} ( 1 - f(g_t^j)) > \lambda_{\alpha}.
\label{eq:surface_rendering}
\end{equation}





\section{Online Training of Dynamic GS}
When training a static GS, the conventional approach is to first render RGB \& depth images at target views as usual, and then minimize the loss between the rendered and observed ground truth, as is done in \cite{rtgslam,splatam,gsslam_cvpr1,gsslam0}.
However, this approach is insufficient for real-time training of dynamic GS because of the objects' motion.
Here, we introduce an improved training method that first explicitly modifies the GS to match the current observations to accurately account for object motion. 

\noindent
\label{prerequisites}
\textbf{Prerequisites} We assume that there are suitable submodules for 1) 2D optical flow calculation; 2) motion segmentation; and 3) localization. Although challenging, suitable prior methods exist for these tasks. For 2D optical flow, we adopted the real-time optical flow (RAFT)~\cite{raft}. For motion segmentation, we combined a coarse motion blobs computed from the RAFT optical flow image with real-time online SAM2~\cite{sam2} (Sec. \ref{Dynamic GS Segmentation Flow}).
For localization, we adopt DynoSAM~\cite{dynosam}, as aforementioned in Sec.~\ref{sec:problem_formulation}, and use its estimated camera poses without modification. 

\subsection{Dynamic GS Flow}
To make the most of the current observations, we utilize the optical flow at the current frame $t$ to associate and propagate the existing GS from time $t-1$.
To do so, we lift the 2D optical flow to 3D (Fig. \ref{fig:overview}(a)), and call it \emph{dynamic GS flow}.
To obtain the dynamic GS flow in 3D from the current to last frame $t \rightarrow t-1$, we mask out static optical flow with 2D motion mask $M_t$, project the moving optical flow $\mathbf{f}_{t-1 \gets t}(u,v)$ to 3D dynamic GS flow $\mathbf{F}_{t-1 \gets t}^{\text{Dyna}}$ using depth $D_t$, and compensate the ego motion: 
\begin{equation}\label{eq:gs_flow}
\mathbf{F}_{t-1 \gets t}^{\text{Dyna}} = M_t \cdot \left( D_{t} K^{-1} \mathbf{f}_{t-1 \gets t} \right) - \left( T_{t-1 \gets t} \mathbf{P}_t - \mathbf{P}_t \right), 
\end{equation}
where $K$ is the camera intrinsic, $P$ is the camera pose,
and $T_{t-1 \gets t}$ is the ego motion transformation


\subsection{Dynamic GS Management}
\label{Dynamic GS Management}
Maintaining an appropriate number of GS is important for GS-SLAM, and even more so in the presence of dynamic objects.
Adding new GS allows capturing the latest information, whereas adding too many will lead to prohibitively high memory usage (n.b. results of SplaTAM \cite{splatam} on ``rpy'' in Table \ref{table:tum}).
Similarly, deleting old GS is essential for capping the memory usage, and more importantly to avoid introducing outliers in the future when it is outdated (n.b. result of RTGSLAM \cite{rtgslam} in Fig. \ref{fig:fig2}).

In this work, we present a management strategy for addition and deletion of the dynamic GS. 
We store dynamic and static GS separately, with different addition and deletion strategies, although they are rendered jointly.

For static GS, we follow the strategy of SOTA static GS-SLAM \cite{rtgslam,splatam}. This strategy adds and optimizes new GS when new areas are seen, while keeping the old GS unchanged. If a GS is not seen for a certain number of frames, it is deleted.


For dynamic GS, this strategy is insufficient because of the objects' motion.
We thus introduce a novel dynamic GS management algorithm shown in Fig. \ref{fig:overview}(b) that overcomes the limitations above.
Let $\mathcal{G}_{t-1} = \{g_{t-1}^{i}\}$ be the GS up to time $t-1$, and let $\mathcal{P}_{t} = \{p_{t}^{j}\}$ be the current motion-segmented RGBD pointcloud.
We first transform the current pointcloud $\mathcal{P}_{t}$ (red) to the previous timestep $t-1$ (blue), using the GS flow $\mathbf{F}_{t-1 \gets t}^{\text{Dyna}}$ (green)~\eqref{eq:gs_flow}.
Transforming the current pointcloud back in time using the latest optical flow observation $\mathbf{F}_{t-1 \gets t}^{\text{Dyna}}$ is better than using the cubic Hermite spline~\eqref{eq:cubic_hermite} for extrapolation, because the cubic Hermite spline only contains information up to time $t-1$.
Moreover, by filtering the GS at time $t-1$, we avoid unnecessary foward-propagation of unnecessary GS to time $t$.

With the transformed pointcloud $\mathbf{F}_{t-1 \gets t}^{\text{Dyna}}\mathcal{P}_{t}$ (green), we search for the nearest neighbor in the existing dynamic GS (blue), and compare the nearest neighbor distance $d_{\min}(p_{t}^i)$ the average nearest neighbor distance $\bar{d}$, computed as:
\begin{equation}
\bar{d} = \frac{1}{N_W} \sum_{i=1}^{N_W} d_{\min}(p_t^i) = \frac{1}{N_W} \sum_{i=1}^{N_W} \min_{g \in \mathcal{G}_t} \| \mathbf{F}_{t-1 \gets t}^{\text{Dyna}} p_t^i - g\|.
\end{equation}
We check if the nearest neighbor distance exceeds some ratio threshold $\lambda_{d}$ of $\bar{d}$, resulting in two cases: 



\noindent
\textbf{Case 1 (prev. observed points):} 
$d_{\min}(p_{t}^{i}) \leq \lambda_{d} \bar{d}$. 
The nearest past GS (blue) is within the distance threshold of the transformed point (green). In this case, we simply reuse the past GS (blue), by replacing their mean with the current point matched (red), and optimize using the current RGBD.
This explicit modification is the key to higher performance, as it moves the past GS to the right location in one step, whereas the usual gradient updates as in the static GS case can only provide minor, insufficient displacements.

\noindent
\textbf{Case 2 (new points)}: $d_{\min}(p_{t}^{i}) > \lambda_{d} \bar{d}$. There is no past GS (blue) near the transformed point (green). In this case, a new GS (red) is initialized for the point $p_t^i$. This allows complete coverage of the whole scene, as some objects are unseen before the current frame (e.g. occluded sides of the moving box in Fig. \ref{fig:fig2}).

We also check the validity of the existing GS against two conditions: observability and longevity.

\noindent
\textbf{Cond. 1 (observability):} $\exists p^i_t \in \mathcal{P}_{t}, \ \|\mathbf{F}_{t-1 \gets t}^{\text{Dyna}}p_t^i - g_j\| \leq \lambda_{d}\bar{d}$. We only keep GS (blue) that are within distance threshold of the currently observed points (green). This step is essential for dynamic GS because old, unobserved GS become outlier noise in the future if they are not displaced or deleted, unlike in static GS where the scene is unchanged.


\noindent
\textbf{Cond. 2 (longevity):} We delete any GS that persisted for a longer period of time than a set longevity threshold.




As observed, the distance threshold ratio $\lambda_{d}$ plays a crucial role in the dynamic GS management.
We thoroughly study its impact in Sec. \ref{ablation_study}. 

Point trackers may seem to serve the same purpose, but they do not, let alone being too slow for online SLAM (n.b, \cite[Table 1]{DOT}). We are searching for correspondences from the current pointcloud to the past GS, whereas point trackers predict the location of past points in the current frame, necessitating the management logic we presented.


\subsection{Rendering and Optimization}
\label{Dynamic GS Segmentation Flow}
Although being managed separately, we jointly render dynamic \& static GS because it improves their interactions to better handle occlusions, lighting consistency, and spatial coherence. We follow the SOTA GS-SLAM \cite{rtgslam,splatam,gsslam0,gsslam_cvpr1,gsslam_cvpr2} with similar supervision between the rendered (\eqref{eq:volumetric_rendering}, \eqref{eq:surface_rendering}) and input RGBD over a small time window of past views.

Although jointly rendered, the optimization is decoupled in that different learning rates are used between dynamic and static GS, as well as longevity windows.  Static GS attributes remain more stable across frames, while dynamic GS undergo abrupt changes, decoupling ensures that their updates do not interfere with the optimization of static structures. Without decoupling, motion inconsistencies from dynamic objects introduce ghosting effects, blending, or incorrect updates in the static region (as shown in Fig. \ref{fig:fig2}), where remnants of dynamic objects appear in static regions due to incorrect optimization of photometric and geometric consistency.



\noindent
\textbf{Optimization-free update of motion spline.} 
\label{Dynamic GS Motion Tracking Prediction}
After associating and training the dynamic GS with respect to the current frame, the cubic Hermite spline \eqref{eq:cubic_hermite} can be updated analytically without optimization. 
This is because the parameters $\mathbf{m}^i_{t-,t+}$ and $\mathbf{v}^i_{t-,t+}$ in \eqref{eq:cubic_hermite} correspond exactly to the 3D position and the velocity of the center of GS $g^i_t$ at the last ($t-$) and current ($t+$) frame.
Thus, we directly set $\mathbf{m}^i_{t-,t+}$ from the optimized GS center at the last ($t-$) and current ($t+$) frames.
The velocity term at the last frame $\mathbf{v}^i_{t-}$ is set as the negative of the GS flow~\eqref{eq:gs_flow}, so that $\mathbf{v}^i_{t-} = -\mathbf{F}_{ (t-1) \gets t}^{\text{Dyna}}$.
The velocity term at the current frame $\mathbf{v}^i_{t+}$ is estimated using the constant acceleration assumption, by extrapolating between $\mathbf{v}^i_{(t-1)+}$ and $\mathbf{v}^i_{t-}$.
If $\mathbf{v}^i_{(t-1)+}$ is unavailable (e.g. when the GS $g^i_{t}$ was just initialized), we fall back to the constant velocity assumption.

\section{Experiments}

\begin{figure*}[!t]
    \centering
    \includegraphics[width=\linewidth]{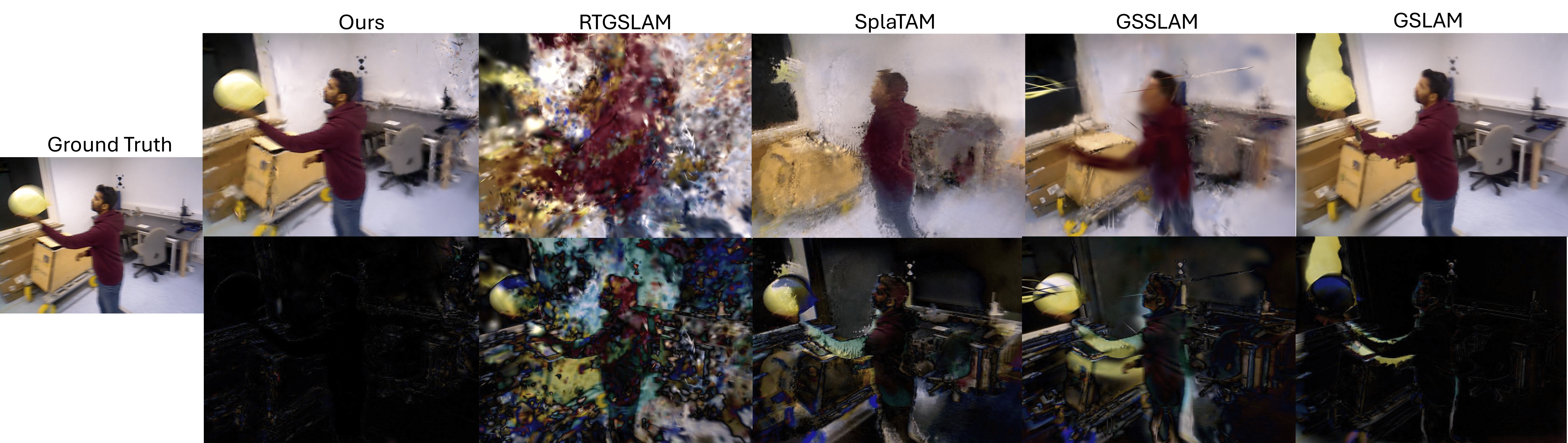}
    \vspace{-5mm}
    \caption{\textbf{Qualitative Results on Bonn Dataset.} First row: RGB rendering. Second row: Difference between the rendering and  the ground truth. Our DynaGSLAM outperforms all baseline static GS-SLAM on the mapping quality, especially at the moving balloon and person.} 
    \label{fig:bonn_vis_balloon2}
\end{figure*}

\noindent
\textbf{Baselines:} We compare the performance of our DynaGSLAM algorithm against four other SOTA GS-SLAM methods, which are: RTGSLAM \cite{rtgslam}, SplatTAM \cite{splatam}, GSSLAM \cite{gsslam_cvpr1} and GSLAM \cite{gsslam0}.
Although there are other concurrent "anti" dynamic GS-SLAM methods~\cite{dgslam,dgsslam,garadslam,gassidy,pgslam} that remove dynamic objects, we could not reproduce these methods because their code is unavailable.
We compare against the reported results where possible. 

\noindent
\textbf{Datasets:} Prior works on static GS-SLAM are evaluated on synthetic datasets with static scenes \cite{replica,scannet}. In contrast, we evaluate our method on real datasets with dynamic scenes. We use OMD \cite{omd} and the dynamic scenes from the TUM \cite{tum} and the BONN \cite{Bonn} datasets. 

\noindent
\textbf{Experimental Setup:}
For TUM and OMD datasets, we use DepthAnythingV2 \cite{depth_anything_v2} to get smooth depth and recover the real scale with the original depth map because the raw depth sensor measurements come with large portion of invalid regions in these datasets. 
For the Bonn dataset, we use the raw depth sensors measurements. 
In addition to the common metrics (PSNR, SSIM, LPIPS) used for mapping evaluation, we evaluate ``DynaPSNR'' as PSNR only for dynamic objects within 2D motion masks. 
We evaluate the Absolute Trajectory Error (ATE) of camera localization.
The experiments are conducted on a desktop with a single NVIDIA 3090Ti (24GB).

\begin{table}[ht]
\centering
\begin{adjustbox}{max width=\columnwidth}
\begin{tabular}{cccccccc}
Metrics         & Scene              & \cite{rtgslam} & \cite{gsslam_cvpr1} & \cite{splatam} & \cite{gsslam0} & \cite{gassidy}$^*$ & Ours \\ \hline
                & \textit{balloon}   & 13.7           & 19.3                & 18.8           & 24.5           & 24.0           & \bf{28.4}      \\
                & \textit{balloon2}  & 12.9           & 17.8                & 16.4           & 23.1           & 22.9           & \bf{28.3}      \\
PSNR$\uparrow$  & \textit{ps\_track} & 13.2           & 14.9                & 15.6           & 24.7           & 24.6           & \bf{28.0}      \\
                & \textit{ps\_track2}& 13.5           & 15.9                & 13.7           & 24.6           & 24.2           & \bf{27.4}      \\ \hline
                & \textit{balloon}   & 38.4           & 73.0                & 73.5           & 85.8           & 77.5           & \bf{93.1}      \\
                & \textit{ballon2}   & 32.0           & 67.5                & 60.6           & 83.2           & 71.5           & \bf{93.3}      \\
SSIM$\uparrow$  & \textit{ps\_track} & 36.1           & 59.2                &54.9            & 86.4           & 78.7           & \bf{93.0}      \\
                & \textit{ps\_track2}& 38.1           & 69.9                &46.9            & 86.2           & 77.3           & \bf{91.6}      \\ \hline
                & \textit{balloon}   & 67.9           & 43.9                &38.8            & 26.6           & 32.5           & 29.3      \\
                & \textit{balloon2}  & 71.2           & 46.2                &51.2            & 27.2           & 39.4           & \bf{26.6}      \\
LPIPS$\downarrow$ & \textit{ps\_track}& 69.2          & 54.8                &53.4            & 24.3           & 32.8           & 28.9      \\
                & \textit{ps\_track2}& 66.5           & 45.4                &56.2            & 24.3           & 32.0           & 31.6      \\ \hline
                & \textit{balloon}   & 18.8           & 14.6                &15.2            & 19.8           & -              & \bf{32.5}      \\
                & \textit{balloon2}  & 16.6           & 14.1                &14.6            & 20.0           & -              & \bf{32.8}      \\
DynaPSNR$\uparrow$ & \textit{ps\_track}& 18.1         & 8.8                 &10.3            & 21.6           & -              & \bf{32.1}      \\
                & \textit{ps\_track2} & 17.4          & 7.6                 &8.7             & 22.2           & -              & \bf{32.6}      \\      
\end{tabular}
\end{adjustbox}
\caption{\textbf{Comparison on Bonn Dataset.} Our method outperforms all other baselines. $^*$: Reported results from \cite{gassidy} listed without reproduction due to unavailability of code. DynaPSNR unavailable for \cite{gassidy}, because dynamic objects are removed.}
\label{table:bonn}
\end{table}

\begin{table}[ht]
\centering
\begin{adjustbox}{max width=\columnwidth}
\begin{tabular}{cccccccc}
Metrics         & Scene       &\cite{rtgslam} & \cite{splatam}  & \cite{gsslam_cvpr1} & \cite{gsslam0}  & Ours \\ \hline
                & \textit{fr3\_wk\_xyz}     & 14.3  & 13.8  & 13.5  &22.0  & \bf{27.5}  \\
PSNR$\uparrow$  & \textit{fr3\_wk\_static}  & 13.9  & 15.5  & 15.9  &20.2  & \bf{26.9}  \\
                & \textit{fr3\_wk\_rpy}     & 15.2  & OOM   & 13.7  &25.0  & \bf{27.4}  \\
                & \textit{fr3\_wk\_hs}      & 13.1  & 11.9  & 12.3  &24.7  & \bf{27.2}  \\ \hline
                & \textit{fr3\_wk\_xyz}     & 45.4  &40.8   & 38.2  &80.7   & \bf{95.7}         \\
SSIM$\uparrow$  & \textit{fr3\_wk\_static}  & 52.9  &60.5   & 54.7  &73.5   & \bf{96.1}  \\
                & \textit{fr3\_wk\_rpy}     & 51.7  & OOM   & 40.3  &88.3   & \bf{94.7}  \\
                & \textit{fr3\_wk\_hs}      & 34.3  & 33.4  & 37.0  &87.9   & \bf{94.5}  \\ \hline
                & \textit{fr3\_wk\_xyz}     & 59.7  & 64.1  & 58.0  &23.7  & \bf{16.0}         \\
LPIPS$\downarrow$  & \textit{fr3\_wk\_static} &53.9 & 41.7  & 42.4  &29.4  & \bf{14.0}  \\
                & \textit{fr3\_wk\_rpy}     & 56.7  & OOM    & 53.9 &16.8   & 21.1  \\
                & \textit{fr3\_wk\_hs}      & 69.9  & 67.3  & 66.8  &16.3   & 20.0   \\ \hline
                & \textit{fr3\_wk\_xyz}     & 17.0  &12.3   & 12.9  &23.5   & \bf{31.5}        \\
DynaPSNR$\uparrow$ & \textit{fr3\_wk\_static} &16.6 &12.1   & 12.7  &22.5   & \bf{30.3}  \\
                & \textit{fr3\_wk\_rpy}     & 16.5  & OOM     & 12.8  &26.1   & \bf{30.1}  \\
                & \textit{fr3\_wk\_hs}      & 14.9  &12.1   & 12.7  &26.2   & \bf{30.7}
\end{tabular}
\end{adjustbox}
\caption{\textbf{Comparison on TUM Dataset.} Our method outperforms others on all metrics. Best results boldfaced. OOM indicates out of memory.}
\label{table:tum}
\end{table}

\noindent
\textbf{Dynamic Mapping Results.} 
Tables \ref{table:bonn} and \ref{table:tum} show the quantitative comparison of the GS mapping, our DynaGSLAM achieves superior results that outperforms other SOTA GS-SLAM on all dynamic sequences. The superior results in DynaPSNR illustrates the efficacy of our dynamic GS management algorithm. Figures \ref{fig:fig1} and \ref{fig:bonn_vis_balloon2} show some qualitative comparisons, where our rendering quality is better than other works, especially around dynamic objects such as the two people (Fig. \ref{fig:fig1}) and the balloon (Fig. \ref{fig:bonn_vis_balloon2}). \textbf{Please confer the videos in Supplementary Material for full comparison.} Our rendering results exhibit some minor ``floater'' artifacts, because we use very few numbers of GS for efficiency compared to others (Table \ref{table:speed_memory}).
In contrast, SplaTAM \cite{splatam} runs out-of-memory(OOM) after 500 frames because it fails to delete outlier dynamic GS and release memory.

\begin{figure*}[t]
    \centering
    \includegraphics[width=\linewidth]{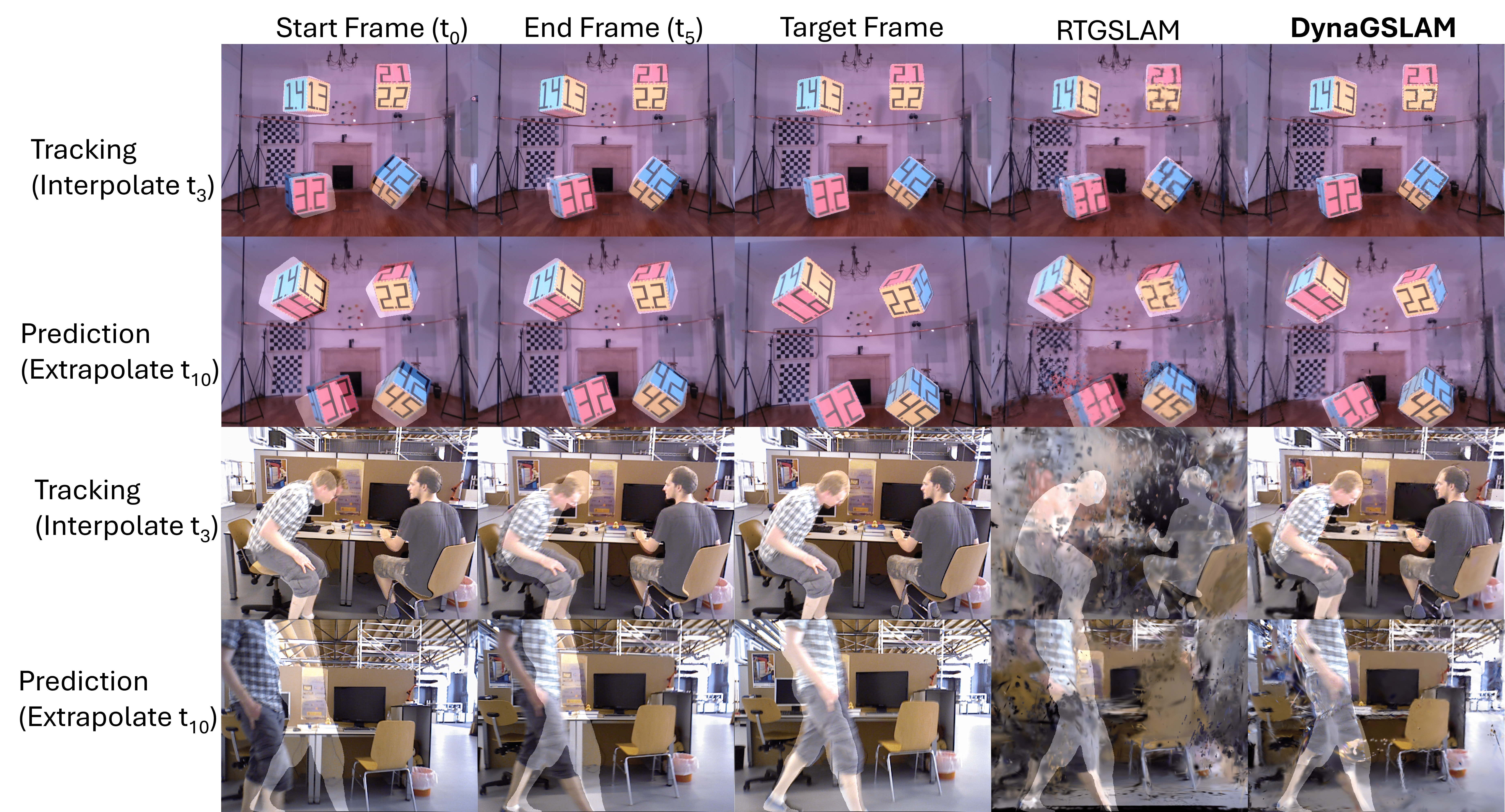}
    \vspace{-2mm}
    \caption{\textbf{Tracking \& Prediction results on OMD and TUM dataset.} ``Start/End Frame'' denotes two consecutive frames ($t_{0}$ \& $t_{5}$). ``Target Frame'' is the ground truth of $t_{3}$ (for tracking) and $t_{10}$ (for prediction). Only ``Start/End Frame'' can be seen by SLAM, ``Target Frame'' cannot be seen by SLAM. When DynaGSLAM online proceeds to the "End Frame" ($t_{5}$) as the current frame, it interpolates the missing past frame $t_{3}$ and extrapolates the unseen future frame $t_{10}$. To better visualize motion quality, we overlap the ground-truth motion mask (white transparent) of the target frames to all frames of ground truths and estimations; A better overlapping between the moving objects and the masks indicates better motion estimation and rendering. Please zoom in to check details.} 
    \label{fig:track_prediction}
\end{figure*}
\noindent
\textbf{Dynamic Motion Tracking \& Prediction.} Figure \ref{fig:track_prediction} shows a qualitative comparison of tracking and prediction with DynaGSLAM. 
Tracking is evaluated by interpolating and rendering an intermediate target timestamp ($t_3$) given two start and end frames ($t_0$ and $t_5$). For prediction, we extrapolate and render a future ($t_{10}$) timestamp given the same input. 
This is an extremely difficult task, as we are only given one out of every five frames, which is temporally sparse, to reconstruct the 3rd and 10th frames that are unseen. 
Since previous methods do not model dynamic objects' motion in GS, we take RTGSLAM \cite{rtgslam} as baseline, assuming no moving entities at $t_5$ and  render at the target viewpoint. 

The results in Fig. \ref{fig:track_prediction} shows that our method accurately predicts the moving objects (moving boxes and people), which overlap with the motion mask (transparent white).
A quantitative comparison of tracking \& prediction is provided in the supplementary. 

\begin{figure}[t]
    \centering
    \includegraphics[width=\linewidth]{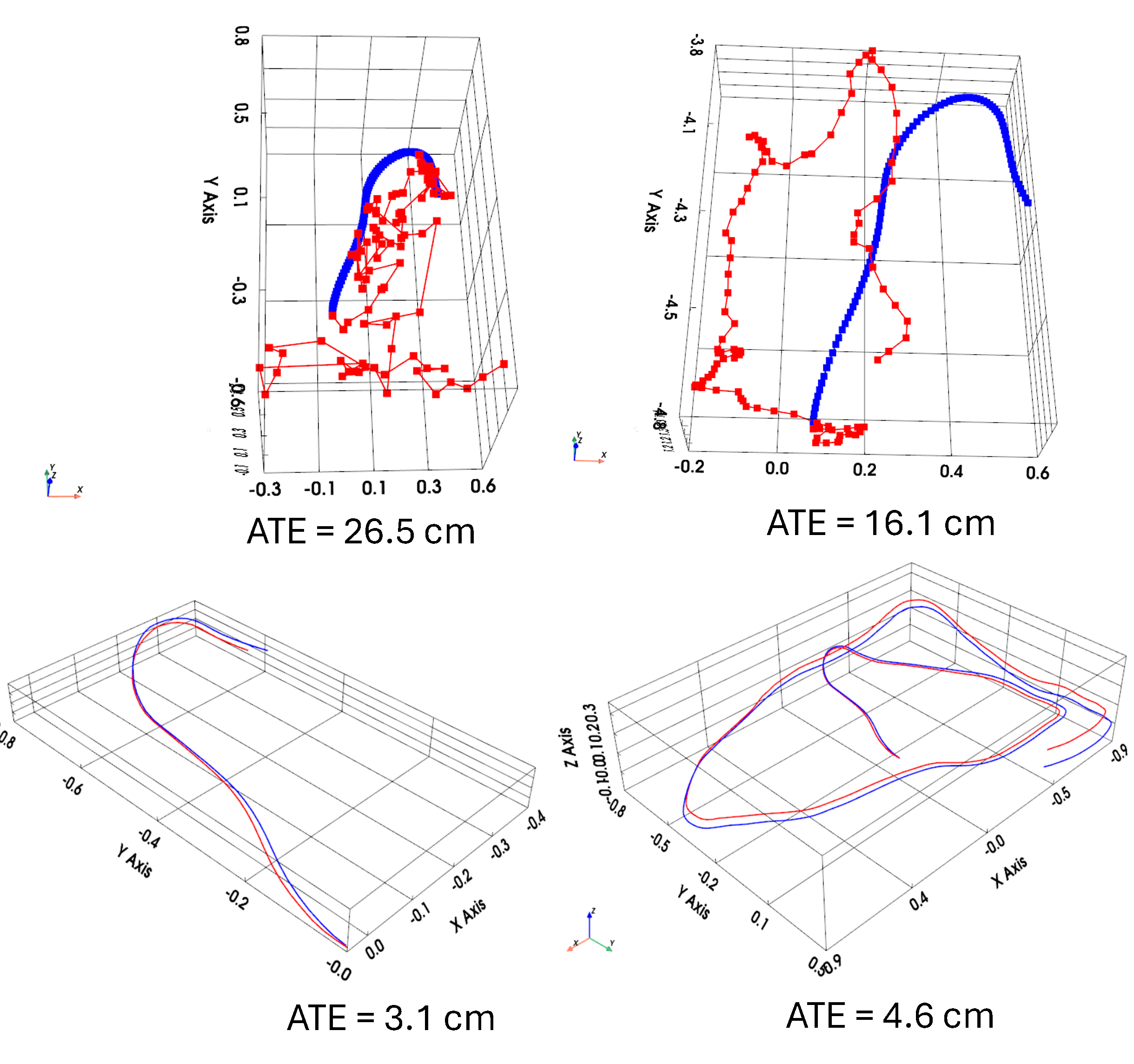}
    \vspace{-5mm}
    \caption{\textbf{Camera Tracking Results} on OMD dataset (S4U). Top Left: RTGSLAM \cite{rtgslam}'s ICP, 90 frames. Top Right: RTGSLAM's ICP refined ORBSLAM2 \cite{orbslam2}, 90 frames. Bottom Left: Ours, 90 frames. Bottom Right: Ours, 500 frames.} 
    \label{fig:localization}
\end{figure}
\noindent
\textbf{Localization Results.} Figure \ref{fig:localization} visualizes the camera trajectory and ATE on OMD dataset (S4U). We directly adopt the world-centric graph optimization strategy from \cite{dynosam}, which considers the moving objects. 
In contrast, ICP and ORBSLAM2 used in RTGSLAM do not distinguish between static and dynamic objects. The result validates the importance of static/dynamic separation. 

\begin{table}[]
\begin{tabular}{cccc}
Methods                                                                            & SplaTAM \cite{splatam} & RTGSLAM \cite{rtgslam} & Ours \\ \hline
\multicolumn{1}{c|}{\begin{tabular}[c]{@{}c@{}}Mapping\\ (ms/frame)\end{tabular}} & 1027       & 555         &347      \\ \hline
\multicolumn{1}{c|}{\begin{tabular}[c]{@{}c@{}}Localization\\ (ms/frame)\end{tabular}}  & 5179         &59         &95     \\ \hline
\multicolumn{1}{c|}{GS number}                                                     &310K         & 520K        &  22K    \\ \hline
\multicolumn{1}{c|}{\begin{tabular}[c]{@{}c@{}}Memory\\ (GB)\end{tabular}}         & 0.82        & 1.7       &  2.6   
\end{tabular}
\caption{\textbf{Comparison of Inference Speed \& Memory} on TUM fr3\_walking\_xyz with a single 3090Ti GPU.}
\label{table:speed_memory}
\end{table}
\noindent
\textbf{Online Speed \& Memory.} Computational speed and memory usage are important for online SLAM. We compare online speed and memory usage on TUM ``fr3\_wk\_xyz'' sequence. The results are shown in Table \ref{table:speed_memory}. For SplaTAM \cite{splatam}, we follow their official configurations for TUM with 200 iters/frame for mapping optimization. 
For both RTGSLAM \cite{rtgslam} and our DynaGSLAM, we follow RTGSLAM's official configuration with 50 iters/frame for mapping optimization. 
SplatTAM and RTGSLAM only update active GS to reduce the memory usage, but their static GS management fails on dynamic scenes yielding much larger number of GS than ours. 
Thanks to our dynamic GS management strategy, we achieve better mapping with much fewer GS. 
Our mapping runs at $\sim$347 ms/frame, including the online SAM2 segmentation ($\sim$36ms) and RAFT \cite{raft} optical flow ($\sim$55ms). 
Our memory usage is higher than baselines mainly due to online SAM2 \cite{sam2} ($\sim$1100mb) and RAFT ($\sim$980mb), but it is acceptable given the benefits of dynamic rendering over baselines. 
The DynoSAM localzation \cite{dynosam} costs $\sim$450 ms/frame.
Overall, the fast computation guarantees real-time operation with  efficient memory usage, while ensuring accurate mapping and localization.

\begin{table}[]
\begin{tabular}{lllll}
Dist Threshold $\lambda_{d}$                       & 0     & 0.01  & 0.1   & 0.5   \\ \hline
\multicolumn{1}{l|}{PSNR$\uparrow$}            & 30.63 & 29.15 & 26.15 & 23.95 \\
\multicolumn{1}{l|}{DynaPSNR$\uparrow$}        & 34.58 & 28.20 & 20.21 & 17.04 \\
\multicolumn{1}{l|}{SSIM$\uparrow$}            & 95.13 & 93.88 & 90.89 & 87.49 \\
\multicolumn{1}{l|}{LPIPS$\downarrow$}           & 15.13 & 17.51 & 22.88 & 27.50 \\
\multicolumn{1}{l|}{Reuse Rate (\%)$\uparrow$} & 0   & 22.75 & 56.36 & 83.53 \\
\multicolumn{1}{l|}{Dyna GS Num $\downarrow$}     & 48.0k & 42.8k & 38.3k & 25.4k
\end{tabular}
\caption{\textbf{Ablation Study} on the distance threshold of the nearest neighbor in Dynamic GS management. Reuse Rate is the ratio of  dynamic GS that are used for at least two consecutive frames. With higher distance threshold, the GS reuse rate is higher, and less new GS are initialized, yielding lower mapping quality but higher efficiency with lower number of dynamic GS. }
\label{table:dist_thresh}
\end{table}

\noindent\
\label{ablation_study}
\textbf{Ablation Study.} The distance threshold $\lambda_{d}$ is the most important hyper-parameter for our dynamic GS management algorithm (Fig. \ref{fig:overview}(b) \& Section \ref{Dynamic GS Management}). Table \ref{table:dist_thresh} shows the effect of distance threshold on mapping quality and computational efficiency. 
In our experiments, we chose a low distance threshold of $\lambda_d=0.05$ with a limit of 50k dynamic GS for the best mapping quality.
In other applications, it may be beneficial to trade off the mapping quality for computational efficiency.
Ablation studies of other factors are presented in the \textbf{Supplementary Material} due to space.

\section{Conclusion and Future Works}
We build the first online GS-based SLAM system - DynaGSLAM that render, track, and predict the motions of dynamic objects with ego motion estimation in real time. Our experiments on three real datasets validate the high quality, efficiency and robustness of our dynamic Gaussian mapping, along with accurate real-time ego motion estimation. To enable the online real-time usage, we pursue efficiency and sacrifice some complexity of the motion model. Future works should explore to various motion models/functions while keeping the system's efficiency.

{
    \small
    \bibliographystyle{ieeenat_fullname}
    \bibliography{main}
}

\clearpage
\newpage
\setcounter{page}{1}
\maketitlesupplementary
\appendix

\section{Dynamic GS Segmentation \& Flow}
\label{Dynamic GS Segmentation Flow}
While online 3D motion segmentation is challenging and slow today, we propose a novel Dynamic GS segmentation strategy (Fig. \ref{fig:overview}). Our GS is initialized from the point cloud of RGBD,
, so we estimate 2D pixel motion and align them to the GS motion. 
When our SLAM proceeds to current frame $t$, given two consecutive images $C_t-1, C_{t} \in \mathbb{R}^{W \times H \times 3}$, we first use a real-time optical flow model (RAFT \cite{raft}) to estimate a flow image $f_{t-1 \gets t} \in \mathbb{R}^{W \times H \times 2}$, where each pixel stores its own 2D velocity, and the velocities of moving pixels are distinct from the static pixels. We then, compute the gradient of $f_{t-1 \gets t}$, which detect edges and close the shapes to get a coarse motion mask. By setting a click prompt at the object centroid, we use prompt-based model SAM2 \cite{sam2} to segment 2D motion. Our strategy enables an automatic pipeline to segment moving pixels and filter out the static objects
Whereas the strategy handles well in general cases, it relies on robust optical flow. 

Incorrect segmentation of static objects as dynamic does not deprecate GS quality since our dynamic GS management also handles static GS, it only introduces minor extra computation that could be ignored in practice. On the contrary, treating moving objects as static causes problems as classical static GS management cannot manage dynamic GS. However, our experiments show that our DynaGSLAM has tolerance to the low quality of 2D motion segmentations, which is further discussed in Sec \ref{Impact of the Motion Segmentation Quality} and Fig. \ref{fig:failure motionseg}. 

\section{Trick of Dynamic Mapping Results}
A popular un-written trick when evaluating mapping metrics (PSNR, SSIM, LPIPS) in previous works (like \cite{splatam}) is to set all invalid pixels to be 0, where the invalid mask is defined by the invalid regions of the original depth maps. However, the trick is unfair since setting estimation and ground truth pixel values both to 0 significantly benefits all metrics: 0 and 0 is infinitely similar. For a fair evaluation, we cancel the cheating benefits by evaluating without mask, which is why our implementation results of SplaTAM \cite{splatam} in table \ref{table:bonn} is worse than the results proposed in table 3 of \cite{gassidy}. However, even without the boost of the trick, our mapping accuracy is significantly better than SOTA baselines. 

\section{Ablation Studies}
\noindent
\textbf{Robustness to Motion Segmentation Noise.}
\label{Impact of the Motion Segmentation Quality}
\begin{figure}[t]
    \centering
    \includegraphics[width=\linewidth]{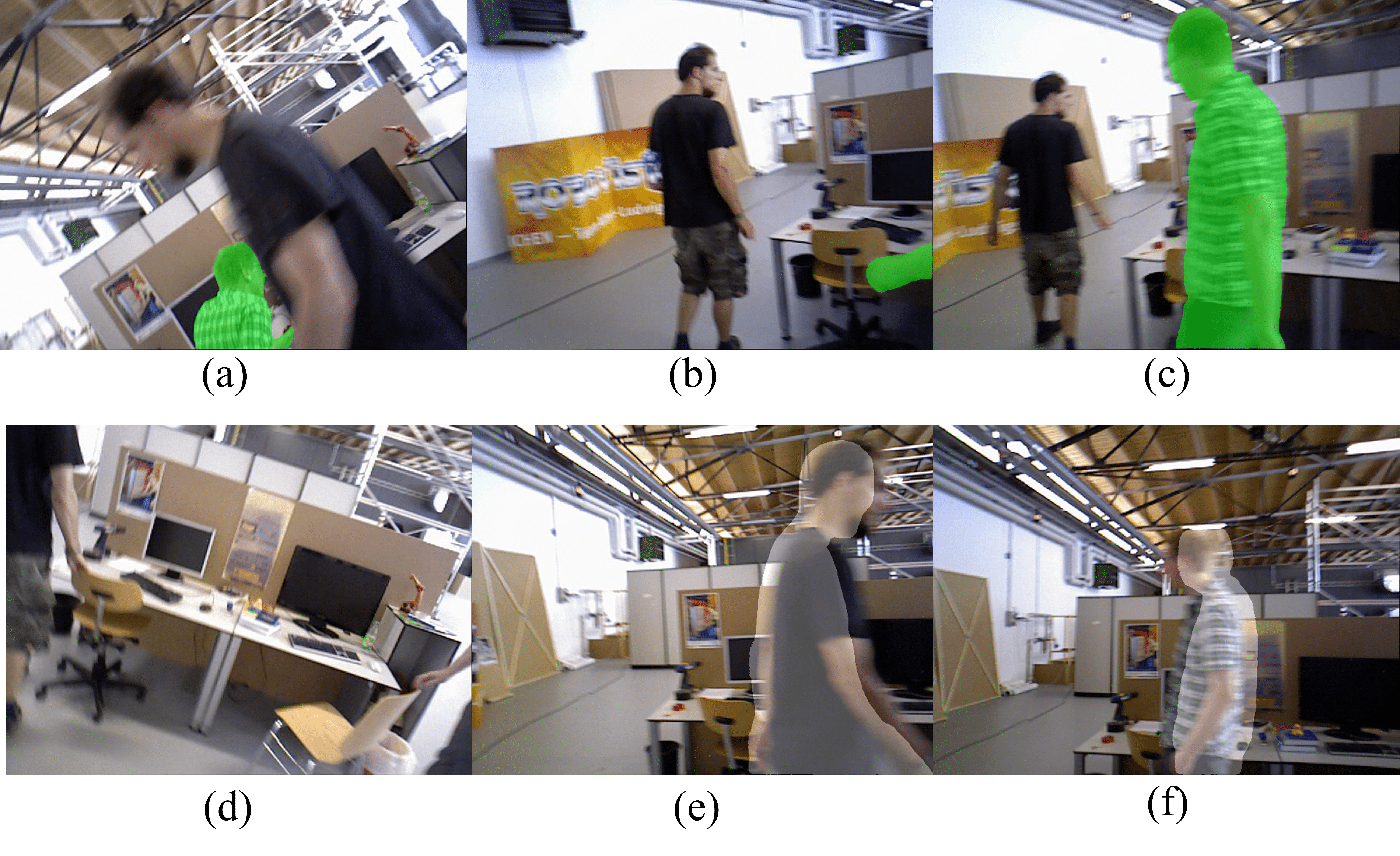}
    \caption{\textbf{Failure of 2D Motion Segmentation} further validates the robustness of our dynamic GS management algorithm under inaccurate motion priors. With some imperfect motion mask,  our DynaGSLAM still enables to reasonably manage dynamic GS, and obtain outstanding mapping quality.} 
\label{fig:failure motionseg}
\end{figure}
As discussed in Sec. \ref{prerequisites} and Sec. \ref{Dynamic GS Segmentation Flow}, 2D motion segmentation is an important prior for our architecture, however, so far there is no perfect solution for online real-time motion segmentation. Although we made improvements on ``automatic'' segmentation, the poor masks can be generated due to: 1. SAM2 \cite{sam2} loses the moving objects in the tracking process, as shown in Fig. \ref{fig:failure motionseg}(abc). 2. The optical flow gradients are not strong enough to initialize any click prompts when the speed of moving object is low, as shown in Fig. \ref{fig:failure motionseg}(d) 3. The SAM2 tracker fails to perfectly segment the fast moving objects due to highly blurred shapes (as shown in \ref{fig:failure motionseg}(ef)). However, with noisy motion masks, our DynaGSLAM still achieves outstanding GS mapping quality, which further validates the robustness of our dynamic GS management algorithm under
inaccurate motion priors.

\begin{table}[t]
\centering
\resizebox{\columnwidth}{!}{%
\begin{tabular}{l|llll}
             & PSNR$\uparrow$      & SSIM$\uparrow$      & LPIPS$\downarrow$     & DynaPSNR$\uparrow$  \\ \hline
OMD (S4U)    & 30.6/31.0 & 95.1/95.7 & 15.1/15.1 & 34.6/31.1 \\
TUM (xyz)    & 27.5/16.3 & 95.7/79.3 & 16.0/37.1 & 31.5/28.7 \\
TUM (static) & 26.9/12.8 & 96.1/77.7 & 14.0/37.8 & 30.3/27.0 \\
TUM (rpy)    & 27.4/20.3 & 94.7/84.6 & 21.1/34.6 & 30.1/29.6 \\
TUM (halfsphere) & 27.2/19.4 & 94.5/83.1 & 20.0/35.4 & 30.7/31.6    
\end{tabular}%
}
\caption{Ablation Study on the Impact of the Depth Quality. In each cell, the metric is ``refined (DepthAnythingV2) depth/original sensor-depth''.}
\label{table:depth_ablation}
\end{table}

\noindent
\textbf{Impact of depth quality.} While our model is not very sensitive to noisy 2D motion priors, it relies on good depth. While the three datasets used in our work are all real datasets with the depth from sensor, TUM's depth is unreliable. We achieved outstanding results on Bonn and OMD with their original depth (table \ref{table:bonn} and \ref{table:depth_ablation}). However, the depth maps from TUM include too large invalid regions, resulting point cloud in poor quality. Nonetheless, we use the original sensor-depth from OMD (table \ref{table:depth_ablation}) and Bonn (table \ref{table:bonn}) to get outstanding mapping quality, which still proves the robustness of our DynaGSLAM with practical depth sensor resource. Moreover, although the PSNR with noisy depth is not ideal, its counterpart ``DynaPSNR'' is still competitive. The good mapping quality of the dynamic region regardless of the static scene further validates our proposed novel dynamic GS management.

\section{Additional Results}
\begin{figure*}[t]
    \centering
    \includegraphics[width=\linewidth]{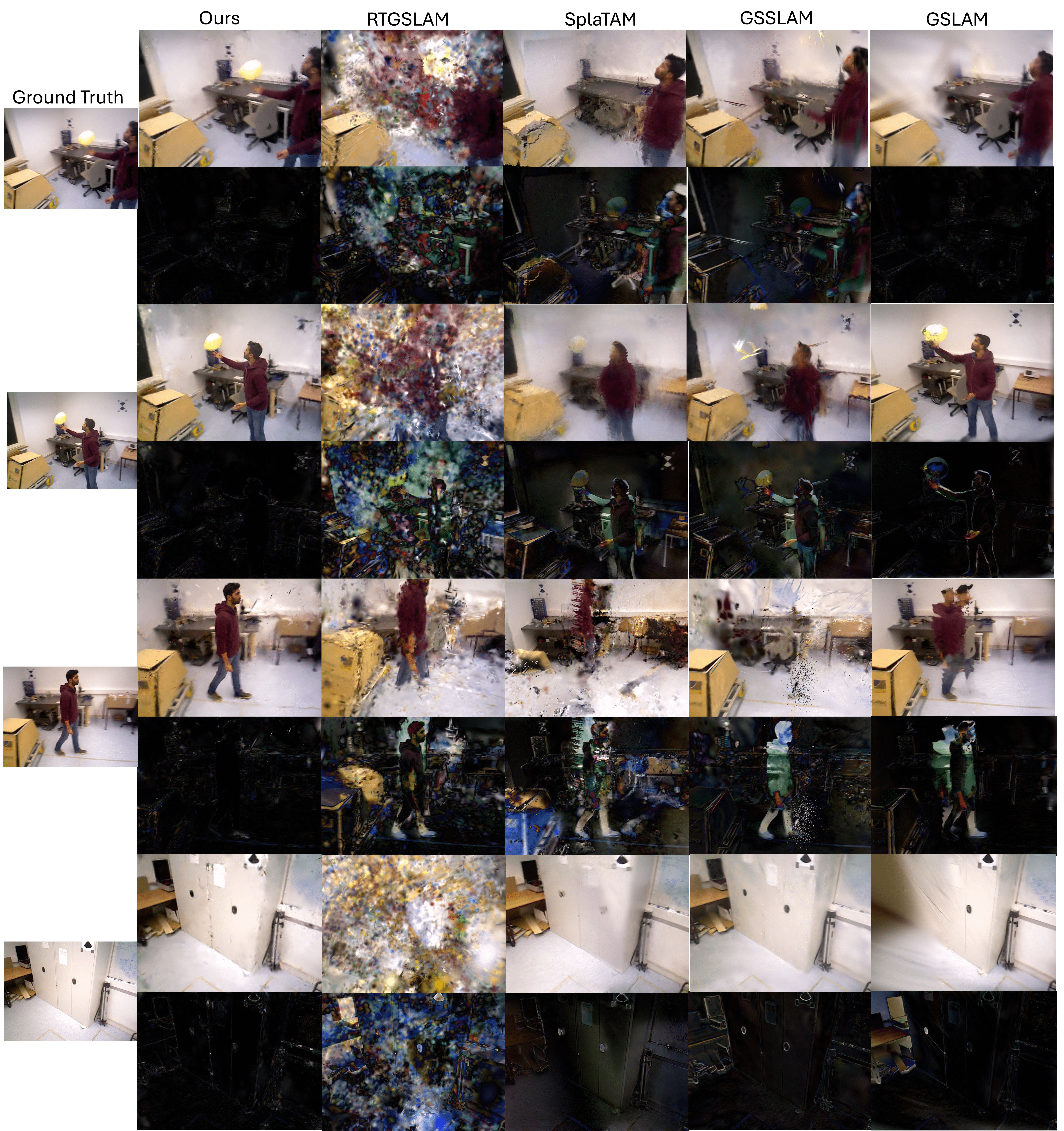}
    \caption{\textbf{GS Mapping Rendering Comparisons on the Bonn Dataset.} From top to bottom, the four scenes are: $balloon$, $balloon2$, $ps\_track$, $ps\_track2$. For each scene, the first row shows the RGB rendering results, the second row shows the absolute error between the rendered RGB to the ground truth. Our DynaGSLAM is obviously better than other SOTA GS-SLAM, especially at the moving entities such as the yellow balloon and the person.} 
    \label{fig:bonn_supp}
\end{figure*}

\begin{figure*}[t]
    \centering
    \includegraphics[width=\linewidth]{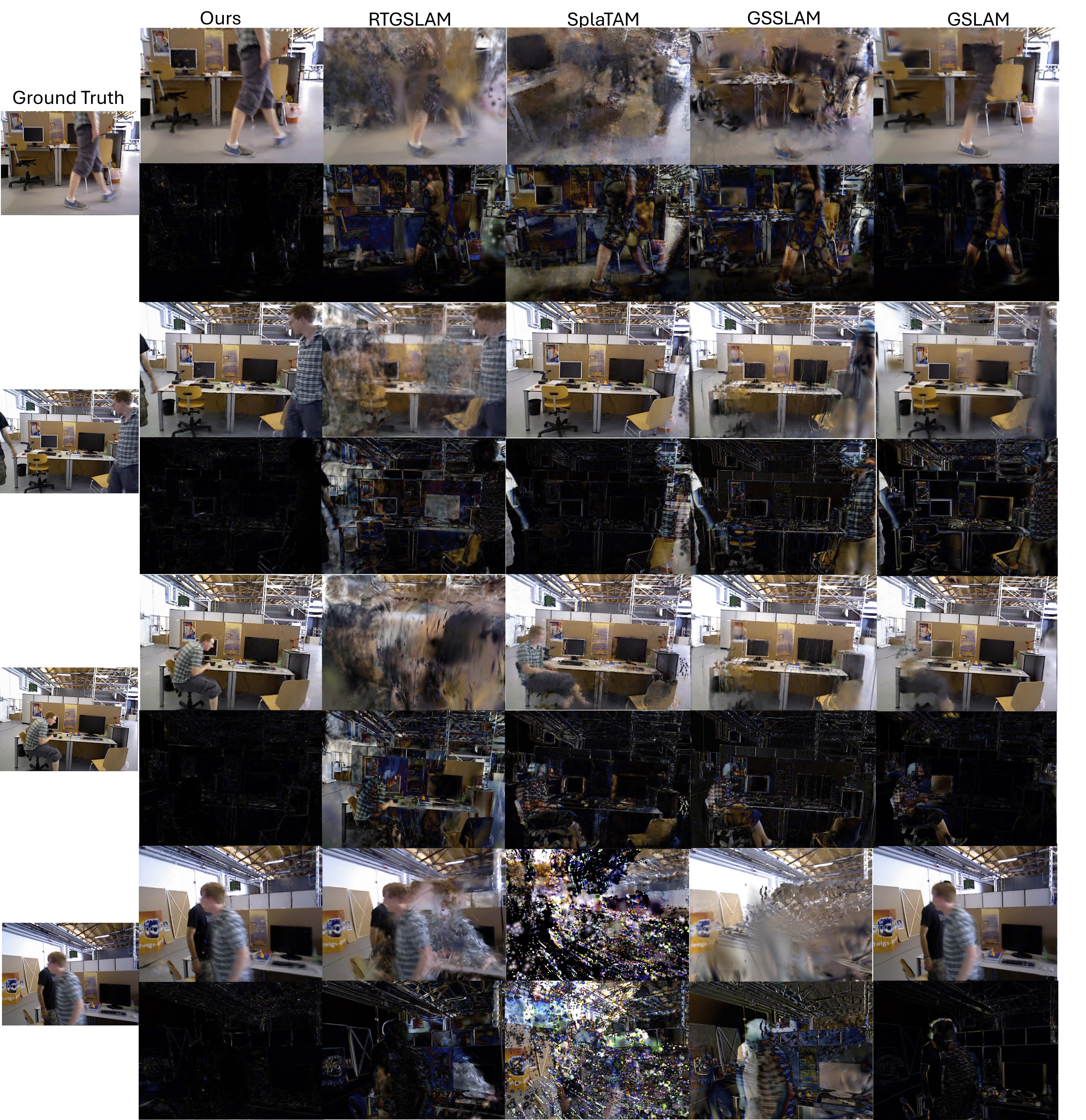}
    \caption{\textbf{GS Mapping Rendering Comparisons on the TUM Dataset.} From top to bottom, the four scenes are: $fr3\_walking\_xyz$, $fr3\_walking\_static$, $fr3\_walking\_static$, $fr3\_walking\_halfsphere$. For each scene, the first row shows the RGB rendering results, the second row shows the absolute error between the rendered RGB to the ground truth. Our DynaGSLAM is obviously better than other SOTA GS-SLAM, especially at the moving people.} 
    \label{fig:tum_supp}
\end{figure*}

\noindent
\textbf{Dynamic Mapping results.} We show additional qualitative comparisons of the GS mapping quality between our DynaGSLAM with SOTA baseline GS-SLAM works (RTGSLAM\cite{rtgslam}, SplaTAM\cite{splatam}, GSSLAM \cite{gsslam_cvpr1}, and GSLAM \cite{gsslam0}). Fig. \ref{fig:bonn_supp} is an extension of Fig. \ref{fig:bonn_vis_balloon2} on the Bonn Dataset. Fig. \ref{fig:tum_supp} is an extension of Fig. \ref{fig:fig1} on the TUM Dataset. Our DynaGSLAM significantly outperforms these baselines, especially around the moving object such as the balloon and moving people. The failures of the baseline GS-SLAM works can be attributed to two aspects: 1. The past dynamic GS cannot be effectively deleted with static GS management, which become outlier GS in the background and contaminate static GS, such as the remnant red GS noises of RTGSLAM in Fig. \ref{fig:bonn_supp}, which belong to the red hoodie of the person in the past frames. 2. The new GS cannot be effectively added with static GS management, such as the missing left leg of GSLAM in \ref{fig:tum_supp} (row 1). Our novel proposed dynamic GS management algorithm overcomes all these limitations proves to be robust and accurate in the real dataset.

\begin{table*}[ht]
\centering
\begin{tabular}{lcccc}
& \multicolumn{2}{c}{\textbf{OMD (S4U)}} 
& \multicolumn{2}{c}{\textbf{TUM (fr3\_walking\_static)}} \\
\cmidrule(lr){2-3}\cmidrule(lr){4-5}
\textbf{Method} & \textbf{DynaGSLAM} & \textbf{RTGSLAM} 
                & \textbf{DynaGSLAM} & \textbf{RTGSLAM} \\
\midrule
\textbf{Mapping}                 & 30.63/34.58 & 17.12/15.66 & 26.88/30.30 & 13.92/16.59 \\
\textbf{Interval = 2 frames, Interpolate middle frame (1st)}  & 20.69/16.69 & 16.90/15.19 & 20.83/19.92 & 14.25/17.07 \\
\textbf{Interval = 5 frames, Interpolate middle frame (3rd)}  & 18.80/14.54 & 16.70/14.46 & 19.98/17.82 & 14.54/16.54 \\
\textbf{Interval = 2 frames, Extrapolate next 1st frame}  & 20.45/16.10 & 16.51/14.55 & 18.64/15.51 & 14.09/16.58 \\
\textbf{Interval = 5 frames, Extrapolate next 1st frame}  & 20.57/16.30 & 17.05/15.32 & 20.01/17.86 & 14.50/16.65 \\
\textbf{Interval = 5 frames, Extrapolate next 2nd frame}  & 18.73/14.33 & 16.44/14.27 & 17.76/14.85 & 14.24/15.82 \\
\textbf{Interval = 5 frames, Extrapolate next 5th frame}  & 17.24/13.30 & 15.65/13.14 & 15.82/14.84 & 13.99/12.27 
\end{tabular}
\caption{\textbf{Ablation Study on Motion Horizon for Tracking and Prediction.} In each cell the metric is represented as ``PSNR$\uparrow$/DynaPSNR$\uparrow$''. The conditions on ``Interval = 5 frames, Interpolate middle frame (3rd)'' and ``Interval = 5 frames, Extrapolate next 5th frame'' are the tracking and prediction corresponding to Fig. \ref{fig:track_prediction} and Fig. \ref{fig:track_prediction_supp}. Please check the annotation explaination in Fig. \ref{fig:track_prediction} to understand the condition in the table. }
\label{table:extrap_interp}
\end{table*}

\begin{figure*}[t]
    \centering
    \includegraphics[width=\linewidth]{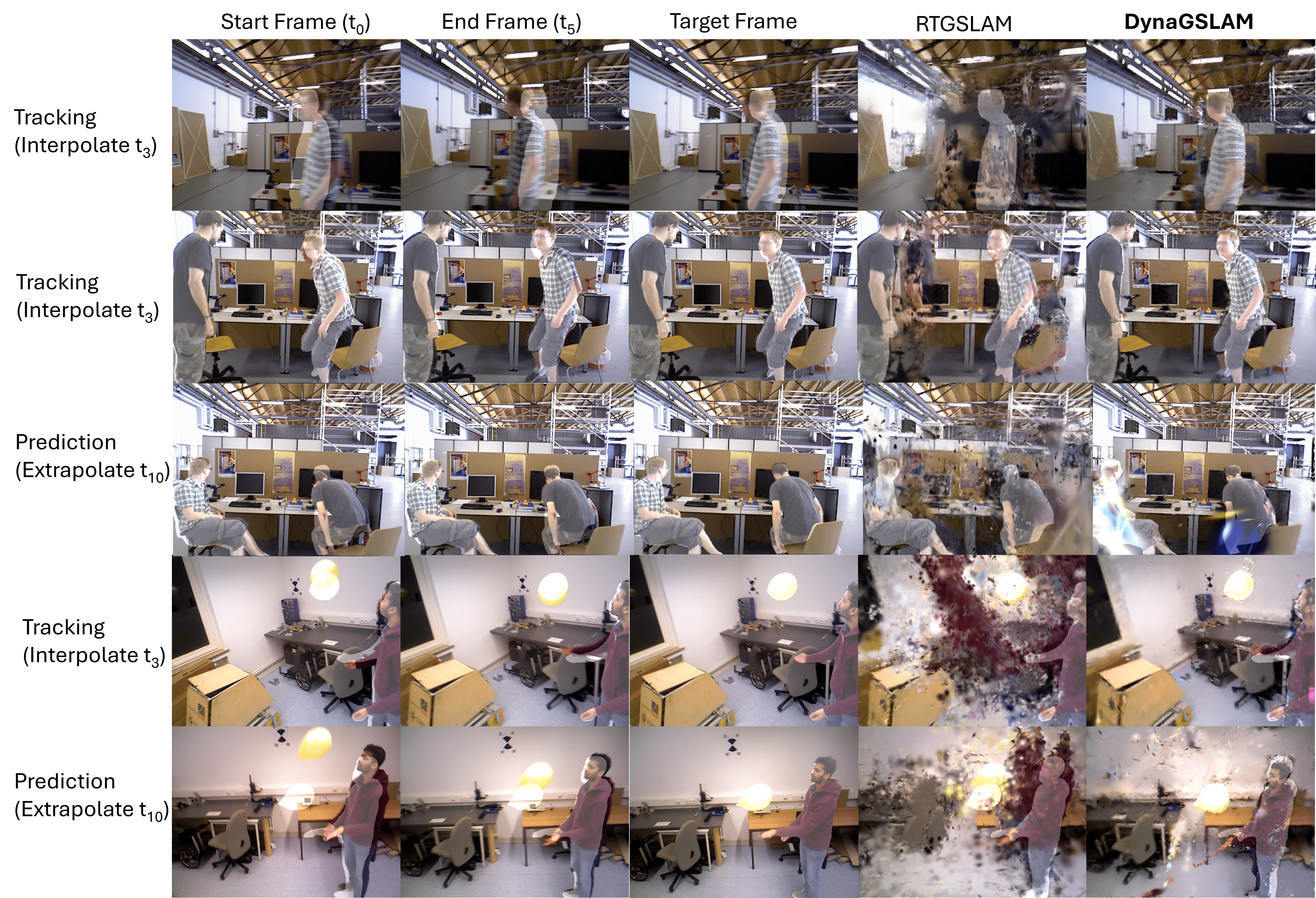}
    \caption{\textbf{Tracking \& Prediction results on OMD, TUM and Bonn datasets.} This figure is an extension of Fig. \ref{fig:track_prediction} showing the tracking and prediction quality of our DynaGSLAM with our proposed novel GS motion function. Please check the annotation explanation in Fig. \ref{fig:track_prediction}. } 
\label{fig:track_prediction_supp}
\end{figure*}

\noindent
\textbf{Dynamic Motion Tracking \& Prediction Results.} We show additional qualitative results of tracking \& prediction in Fig. \ref{fig:track_prediction_supp}. In all of the three datasets, our DynaGSLAM shows the ability to synthesize unseen views by traversing the time dimension. As an extension of Fig. \ref{fig:track_prediction}, we show the tracking (interpolation) and extremely hard prediction (extrapolation) over long missing frames. With the transparent white mask as the ground truth motion, we show that our motion model successfully brings GS to the desirable position, and the overlap of the dynamic entities (balloons and people) with the ground truth motion mask shows the quality of our proposed novel motion function. By contrast, the SOTA static GS-SLAM ``RTGSLAM'' \cite{rtgslam} fails to correct the motion. Due to the lack of dynamic GS management, their background static GS are also contaminated by moving GS. Our DynaGSLAM generates some minor artifacts under the extrapolation of long ``Motion Horizon'', which is mainly because we use an extremely low number of GS for real-time efficiency, so that individual GS can adjust the position and shape to cover more space whereas diminish their photometric textures, this issue can be moderated by trading-off the number and efficiency of GS.

We also conduct quantitative ablation study on the ``Motion Horizon'' for GS tracking and prediction on OMD and TUM datasets, as shown in table \ref{table:extrap_interp}. As the input frame interval or the motion horizon grows, the difficulty for motion estimation is increasing, and tracking(interpolation) always performs better than prediction(extrapolation). While our DynaGSLAM's performance edge is huge under small motion horizon, the advantage gets unclear while the motion horizon is growing. This is because the PSNR metric is very sensitive to even a small displacement - The exact same two patterns get low PSNR if overlapping with a minor displacement. With the motion horizon grows, displacement errors are zoomed out. However, we argue that for a long ``Motion Horizon'', visual results give fairer comparisons, such as the accurate fitting of the moving object's contour with the ground truth motion mask in Fig. \ref{fig:track_prediction_supp}.


\end{document}